%% file: main_arxiv.tex
\documentclass[letterpaper, 10pt, conference]{ieeeconf}
\IEEEoverridecommandlockouts
\usepackage{amsmath,amssymb}
\usepackage{graphicx}
\graphicspath{{figures/}}
\usepackage{booktabs}
\usepackage{multirow}
\usepackage{subcaption}
\usepackage{wrapfig}
\usepackage{xcolor}
\usepackage{cite}
\usepackage{url}
\usepackage{balance}
\usepackage{algorithm}
\usepackage{algpseudocode}
\usepackage[hidelinks]{hyperref}

\newcommand{\method}{PGX}

\title{\LARGE \bf PGX: Physically Grounded Execution for Long-Horizon Humanoid Loco-Manipulation}

\author{Kai Lu$^{1}$, Chenyang Ma$^{2}$, Ioannis Havoutis$^{1}$%
\thanks{$^{1}$K. Lu and I. Havoutis are with the Oxford Robotics Institute, University of Oxford. {\tt\small kai.lu@oxfordrobotics.institute, ioannis@robots.ox.ac.uk}} 
\thanks{$^{2}$C. Ma is with the Department of Computer Science, University of Oxford. {\tt\small chenyang.ma@cs.ox.ac.uk}}}

\begin{document}
\bstctlcite{IEEEexample:BSTcontrol}
\maketitle
\thispagestyle{empty}
\pagestyle{empty}

\begin{abstract}
Robotic vision-language-action (VLA) models can execute diverse individual manipulation skills, yet they still struggle with long-horizon loco-manipulation tasks, where temporal structure, skills, and precision are tightly coupled within a monolithic model. We propose Physically Grounded Execution (PGX), a framework that offloads temporal and precision control to two classical training-free components, allowing the VLA to focus on stage-specific skill dynamics. PGX decomposes long-horizon loco-manipulation into three layers: (i) a predicate-based state machine governs task progression, using physical predicates to gate phase transitions, trigger recovery, and coordinate navigation and waiting. Shared predicate semantics connect offline demonstration segmentation, online execution, and stage-wise evaluation; (ii) VLA skills operate within active manipulation phases, with flat or stage-trained models concentrating on task-relevant actions; and (iii) world-coordinate locking (WCL) maintains spatial consistency through base station keeping, object-relative positioning, and world-anchored contact targets. We demonstrate the effectiveness of PGX with extensive experiments conducted on three simulated tasks and one real-world task using a Unitree G1 humanoid robot for long-horizon loco-manipulation, showing improvements in stage progression and task success rates. Project page is available at \textcolor{blue}{\url{https://kl-research.github.io/pgx}}.



\end{abstract}

\input{intro}

\input{related_work}

\input{method}

\input{experiments}

\input{conclusion}

\balance
\bibliographystyle{IEEEtran}
\bibliography{refs}

\end{document}

%% file: intro.tex

\section{Introduction}

\begin{figure}[t]
  \centering
  \vspace{-3pt}
  \includegraphics[width=\columnwidth]{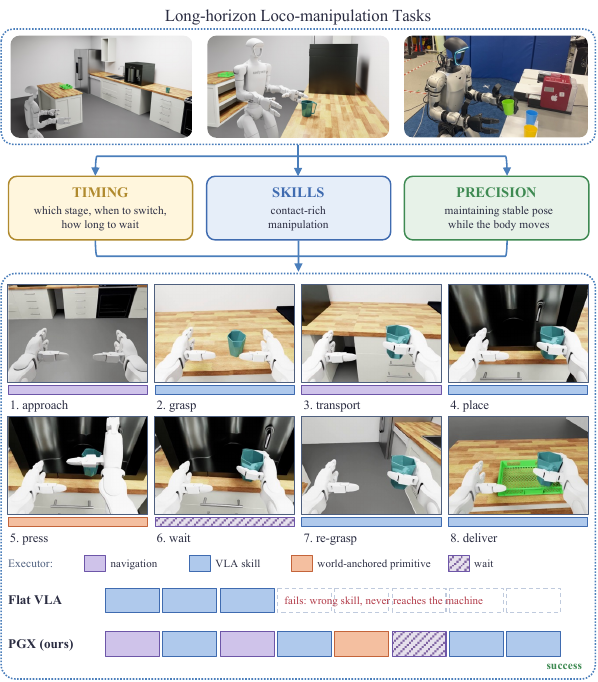}
  \caption{\textbf{Humanoid long-horizon loco-manipulation.} This work studies humanoid loco-manipulation tasks: with the same VLA model, the flat policy fails before the machine, whereas our method \method{} completes the task by keeping timing and world-frame precision outside the network, and the VLA executes only the manipulation stages.}
  \label{fig:teaser}
  \vspace{-6mm}
\end{figure}

Humanoid loco-manipulation for long-horizon tasks is central to embodied intelligence and generalist robotics, with applications in domestic assistance, healthcare, and industrial logistics. Yet even seemingly routine tasks in these domains demand complex, extended coordination between locomotion and manipulation. For instance, fetching a coffee or clearing a table interleaves walking, contact-rich manipulation, and precise object-part interaction over long action sequences. The challenge lies in jointly addressing three tightly coupled aspects: \emph{temporal structure}, \emph{skill dynamics}, and \emph{precise control}, all of which must be grounded in physical interaction and integrated within a unified visuomotor control framework (see Fig.~\ref{fig:teaser}).


Traditional methods for loco-manipulation separate these concerns across modules: task and motion planning or behavior trees control timing~\cite{garrett2021tamp, colledanchise2018behavior, ciebielski2025tamp}, model-based whole-body control owns precision, and manipulation is realized by hand-designed motion primitives. These systems are transparent and precise, but every skill must be engineered for a specific object and scene, and contact-rich skills remain brittle.


Learning-based methods offer a more generalizable alternative: Vision-Language-Action models (VLAs)~\cite{kim2024openvla, black2024pi0, intelligence2025pi05, bjorck2025groot, ding2025humanoidvla, xue2025leverb} and learned whole-body controllers~\cite{fu2024humanplus, ben2025homie, li2025amo} acquire transferable manipulation skills and locomotion behaviors from data, reducing the need for task- or object-specific programming. However, learning and executing the entire long-horizon loco-manipulation task with a single policy introduces difficulties in maintaining task progress, coordinating reference frames, and handling highly imbalanced demonstrations. Without an explicit representation of task progress, visually similar observations from different stages can lead to incorrect skill execution~\cite{ma2026furniturevla}. Moreover, actions represented relative to the moving pelvis frame make precise manipulation difficult, since the action reference changes as the humanoid moves. Long-horizon demonstrations are also highly imbalanced: walking and low-motion segments dominate the trajectory, which can bias imitation learning toward conservative or near-static actions (e.g., standing still). Thus, these challenges remain major obstacles to reliable end-to-end long-horizon loco-manipulation, despite rapid progress in humanoid whole-body control~\cite{cheng2024exbody, ben2025homie, li2025amo} and VLAs.

To this end, we aim to establish a physically grounded framework bridging robot foundation models (i.e., VLAs) and structured robot planning and control. Our insight is that the aspects a flat VLA handles least reliably, deciding \emph{when} and \emph{where} to act, can often be resolved directly from physical quantities, whereas learning \emph{how} to execute a skill is precisely where VLAs are most effective. PGX therefore uses physical structure to determine task progress, skill initiation, and spatial grounding, while leaving the execution of transferable manipulation skills to the VLA. By decoupling timing and world-frame precision from the learned policy, we hypothesize that the same foundation-model backbone can execute long-horizon loco-manipulation more reliably.

Building on this insight, we introduce \method{} (\textbf{P}hysically \textbf{G}rounded e\textbf{X}ecution), a training-free execution framework that separates deciding \emph{when} and \emph{where} to act from learning \emph{how} to act. Rather than modifying the VLA, introducing a learned planner to infer sub-tasks from images~\cite{ huang2022innermonologue, shi2025hirobot, yuan2025being0}, or folding temporal structure into the policy itself~\cite{fan2025longvla, shi2025memoryvla}, PGX assigns task progression and spatial precision to physically grounded components while preserving the VLA as the manipulation skill executor. Specifically, PGX decomposes long-horizon loco-manipulation into three layers. 1) \emph{Timing}: a predicate-gated finite state machine (FSM) maintains task progress and memory, representing each stage as a tuple of instruction, executor, guard predicate, timeout, and retry policy. 2) \emph{Skills}: the VLA operates only within manipulation stages, while the execution layer re-establishes the starting conditions under which each skill was trained; failed guards trigger retries, and reverse transitions recover from prematurely entered stages. 3) \emph{Precision}: world-coordinate locking (WCL) computes navigation goals from the current object pose, holds the base at its manipulation-entry position, and anchors precise contact targets in the world frame. In this way, physical structure governs task progression and spatial precision, allowing the VLA to focus on transferable manipulation dynamics.

We evaluate \method{} in Isaac Lab Arena on three tasks (box transport, mug placement, and coffee fetching) and in a real-world coffee task using a Unitree G1 Humanoid Robot. We fine-tuned the state-of-the-art foundational VLA models $\pi_{0.5}$ and GR00T N1.6 on collected human demonstrations. On coffee fetching, \method{} improves the task success rate of a fine-tuned VLA from 35\% to 80\% with unchanged weights, and improves stage progression and success on all three tasks for both backbones. The real humanoid G1 experiment further demonstrates the effectiveness of the execution structure.



In summary, the main contributions of our work are:
\begin{itemize}
  \item We propose \method{}, a physically grounded execution framework that incorporates VLA models with a predicate-gated state machine and a world-coordinate locking layer, improving long-horizon capability.
  \item We develop predicate-gated stage sequencing with retry and reverse transitions, together with WCL components, and show that the decoupled layers supply structured robot behaviors and generalize across VLA backbones. 
  \item We evaluate the effectiveness of \method{} on three simulated loco-manipulation tasks and on a real humanoid, where it turns near-failing flat policies into reliable long-horizon execution.
\end{itemize}

%% file: related_work.tex

\section{Related Work}


\subsection{Humanoid Loco-Manipulation}
Recent progress on humanoids builds on learned whole-body controllers that track retargeted human motion or teleoperation commands~\cite{cheng2024exbody, he2024omnih2o, fu2024humanplus, ze2025twist, sleiman2026zest, luo2025sonic}, with decoupled upper-body and locomotion interfaces that expose velocity and joint targets to a higher-level policy~\cite{ben2025homie, li2025amo}. On top of such controllers, generalist VLAs~\cite{kim2024openvla, black2024pi0, intelligence2025pi05, bjorck2025groot, nvidia2025groot16} and humanoid-specific policies~\cite{ding2025humanoidvla, xue2025leverb, hu2026openhlm} learn manipulation and locomotion from demonstrations, 
while data-generation pipelines synthesize demonstrations in
simulation through whole-body planning or procedural motion
generation~\cite{mandlekar2023mimicgen, lin2026humanoidmimicgen,
kim2026legs}.
These works primarily improve the controller, the data, or the skill policy. Alternatively, our work takes a tuned VLA and a whole-body control stack as given and studies how to organize their skills into a long loco-manipulation task.
\subsection{Long-Horizon Manipulation with VLAs}
Many recent works tackle long-horizon tasks by decomposing them into sub-tasks, with the active sub-task inferred by an LLM or VLM planner~\cite{ahn2022saycan, huang2022innermonologue, liang2023code, shi2025hirobot}, by the VLA itself through embodied reasoning~\cite{zawalski2024ecot, intelligence2025pi05}, or by a learned success or failure detector~\cite{du2023successvqa}. A second line strengthens the temporal representation inside the policy, through phase-aware input masking~\cite{fan2025longvla}, explicit memory~\cite{shi2025memoryvla}, or action chunking and flow-based decoders~\cite{chi2023diffusionpolicy, black2024pi0}. A third line chains separately learned skills, regularizing terminal states so that one skill starts where the previous one ends~\cite{lee2021skillchaining, agia2023stap, zhang2023boss, cheng2023league, ma2026furniturevla}. These methods leave task progress, the switching decision, and the initial condition of each skill to a network or a semantic model, whereas \method{} chooses to keep task progress in an external state machine and switch skills on physical events rather than inferred semantics.

\subsection{Structured Robotic Execution and Hybrid Control}

\begin{figure*}[t]
  \centering
  \includegraphics[width=\textwidth]{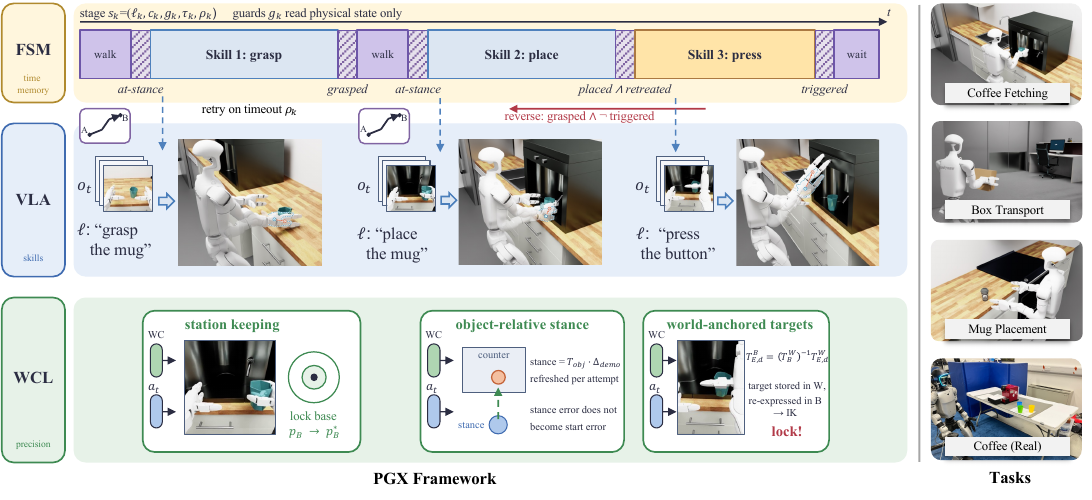}
  \vspace{-5mm}
\caption{\textbf{Overall framework of \method{}.} FSM (top): stages separated by predicate gates, with walking and waiting handled by the executor; retry and reverse edges for failed guards and stage confusion. VLA (middle): within a manipulation stage, the policy maps images and the stage prompt to an action chunk from an executor-prepared start state. WCL (bottom): world-frame references for station keeping, object-relative stance, and contact targets. Right: the loco-manipulation tasks.}
  \label{fig:framework}
\vspace{-3mm}
\end{figure*}

Combining task structure with continuous control has been studied for years, and recent work replaces parts of this pipeline with learned or language-based components. 
Language models now generate the task structure as programs or plans and delegate execution to a library of learned skills~\cite{liang2023code, ahn2022saycan, lu2025kitchenvla}.  ~\cite{yuan2025being0} tracks progress from images with a foundation model and adjusts the stance with a VLM before manipulation, while our approach gates stages on physical predicates rather than image-based inference. Hybrid planners chain learned skills under feasibility checks~\cite{agia2023stap, cheng2023league}.
These build on the classical machinery of task and motion planning~\cite{garrett2021tamp, ciebielski2025tamp} and behavior trees~\cite{colledanchise2018behavior}, which make the discrete structure explicit and hand execution to model-based control.
On humanoids, large learned whole-body controllers now expose one command interface to teleoperation and VLA policies alike~\cite{luo2025sonic, sleiman2026zest}. 
For mobile manipulators, the standing position from which a skill starts is known to dominate its success, motivating coupled base-arm imitation policies~\cite{fu2024mobilealoha}, mostly on fixed or wheeled bases that do not drift during manipulation. 
Our work targets a biped whose stance is less stable than these bases, and builds the interface between learned skills and the legged control stack.



%% file: method.tex

\section{Physically Grounded Execution}

Our goal is to use a hierarchical execution structure to address the key challenges of long-horizon humanoid loco-manipulation, including task progress, spatial precision, and highly imbalanced demonstrations. The overall framework of \method{} is shown in Fig.~\ref{fig:framework}. A long-horizon loco-manipulation task is executed by three layers: a predicate state machine decides which stage is active and who executes it, the VLA acts inside manipulation stages, and a world-coordinate locking (WCL) layer keeps the base, the carried object, and device targets consistent in the world frame. The three layers are a division of responsibility rather than three networks that every action must pass through. This decomposition separates stages with distinct dynamics and learning requirements, preventing long locomotion and low-motion segments from dominating manipulation learning. We state the problem in Sec.~\ref{sec:problem}, describe the three layers in Secs.~\ref{sec:fsm}--\ref{sec:wcl}, and show in Sec.~\ref{sec:skill} how the same predicates segment demonstrations for training.

\subsection{Problem Statement}
\label{sec:problem}
We consider a long-horizon task given as a stage graph $\mathcal{G}=(\mathcal{S},\mathcal{E})$, a trained VLA policy $\pi_\theta$, and a whole-body control stack. The graph contains the nominal progression path and the retry or reverse edges triggered by failure states. The goal is to complete the task within a time budget.


We distinguish the policy observation $o_t$, which contains camera images and proprioception, from the executor state $z_t$, which additionally contains estimates of the base and task-object poses, grasp evidence, and device states, provided by the robot's state estimation. The executor state is used only for gating, navigation, and geometric control. Given $o_t$ and the stage instruction $\ell_k$, the VLA outputs an action chunk of length $H$:
\begin{equation}
  A_t = \pi_\theta(o_t, \ell_k),\qquad A_t = (a_t,\dots,a_{t+H-1}),
  \label{eq:chunk}
\end{equation}
which a model-specific interface maps to upper-body joint targets, base velocity commands, and hand commands, tracked by the whole-body controller. \method{} does not change the network or its parameters $\theta$; at every control period it selects, constrains, or replaces individual action channels according to the stage and the physical state.

\subsection{Predicate-Gated Stage Machine}
\label{sec:fsm}
\textbf{Stage representation.} Each stage is written as a tuple
\begin{equation}
  s_k = (\ell_k,\ c_k,\ g_k,\ \tau_k,\ \rho_k),
  \label{eq:stage}
\end{equation}
denoting its instruction, executor, completion guard, time budget, and recovery policy. Walking stages are executed by a navigation controller while the arms hold a ready or carry posture; manipulation stages are executed by the VLA while the executor manages the base; device waits are handled by a state check or a timer, so the VLA never has to generate a waiting behavior. In this work, the coffee task has eight stages: approach, grasp, transport, place, press, re-grasp, transport, and deliver, of which five are manipulation stages, and the wait between press and re-grasp belongs to the executor. Every stage exits on a physical outcome, never on the end of an action chunk or a preset duration.

\textbf{Predicate interface.} Five predicates form the basic set: \textit{grasped}, \textit{placed}, \textit{triggered}, \textit{at-stance}, and \textit{dropped}. \textit{grasped} combines the hand state, the wrist--object distance, and a lift above the object's rest height,
\begin{equation}
  \textit{grasped} \Leftrightarrow \text{closed} \wedge \|p_{W}-p_{O}\|<d_g \wedge z_{O}-z_{\mathrm{rest}}>h_g,
  \label{eq:grasped}
\end{equation}
so that closing an empty hand is not mistaken for a grasp. \textit{placed} requires the object to be inside the target region, slower than $v_p$, and released; for intermediate placements it also requires the wrist to have retreated, leaving room for the next interaction. \textit{triggered} reads the device state, \textit{at-stance} compares the base pose with the stage's reference stance, and \textit{dropped} fires when a held object leaves the hand outside a placement region. Each predicate must hold for a minimum duration, with hysteresis, so that one noisy frame cannot advance the stage.

The key design choice is that one rule serves three purposes. The same predicate determines skill boundaries in demonstrations offline (Sec.~\ref{sec:skill}), gates transitions online, and records task progress at evaluation. The training target, the execution condition, and the metric are therefore tied to one physical event; each use only sets thresholds and hold times according to its sensor source, and nothing has to redefine what ``grasped'' or ``placed'' means.

\textbf{Recovery and reverse transitions.} Each manipulation stage receives a bounded number of attempts. When the guard fails, the executor restores the stance and requests a fresh action chunk, without releasing a held object and without opening the hand close to the object. For stage confusion the machine allows reverse transitions: if during press the mug is grasped again while the device has not been triggered, the executor returns to place and re-establishes the pre-press condition, changing only the stage and the prompt. An unrecoverable drop terminates the episode. Algorithm~\ref{alg:exec} summarizes the execution of one stage.

\begin{algorithm}[t]
\caption{Execution of one stage $s_k$ in \method{}}
\label{alg:exec}
\begin{algorithmic}[1]
\Require $s_k=(\ell_k,c_k,g_k,\tau_k,\rho_k)$, frozen $\pi_\theta$, executor state $z_t$
\If{$c_k$ is VLA} \State \textsc{Prepare}$(s_k)$ \Comment{stance, still base, arm posture} \EndIf
\While{$\lnot g_k(z_t)$ \textbf{and} $t<\tau_k$}
  \If{$c_k$ is VLA}
    \State $A_t \gets \pi_\theta(o_t,\ell_k)$; act $A_t$, base channel masked
  \Else
    \State \textsc{Step}$(c_k)$ \Comment{navigation or wait}
  \EndIf
  \State \textsc{WCL}$(z_t)$ \Comment{Eqs.~\eqref{eq:sk}, \eqref{eq:press}}
  \If{\textit{dropped}$(z_t)$} \State \Return \textsc{Abort} \EndIf
  \If{timeout \textbf{or} confusion} \State \textsc{Recover}$(\rho_k)$ \Comment{retry or reverse} \EndIf
\EndWhile
\State \Return \textsc{Advance} if $g_k(z_t)$ else \textsc{Fail}
\end{algorithmic}
\end{algorithm}

\subsection{Skill Execution with Fine-Tuned VLAs}
\label{sec:skill}
\method{} treats the VLA as a policy that manipulates under local conditions, not as the sole controller of the whole task. Before a manipulation stage, the navigation controller brings the robot to a stance relative to the target, the executor waits for the stopping transient to decay and selects an arm posture compatible with the coming skill: an empty-hand grasp starts from the ready posture, a placement with an object starts from the carry posture, and consecutive manipulations keep a compatible end-effector state. 

Two kinds of skill model are supported. A \emph{flat model} is trained on full demonstrations with one task-level instruction, kept for every manipulation stage inside \method{}. 
A \emph{stage model} is trained on manipulation segments cut from the same demonstrations by the same predicates that gate execution: a segment starts when effective hand motion begins after the last base adjustment and ends on the grasp, release, or retreat event, keeping the trailing motion that completes the hand-over. Walking and waiting are excluded, so training concentrates on the manipulation itself, and each segment starts from the state the executor re-establishes at deployment. 



\subsection{World-Coordinate Locking}
\label{sec:wcl}
WCL reduces the influence of base motion on skill inputs and on interaction targets by maintaining task-relevant references in the world frame. It acts on the stance and on fine contact separately, rather than locking the end-effector for all manipulation.

\textbf{1) Object-relative stance.} For skills whose start depends on where the object currently is, the navigation goal is expressed relative to the object rather than to the map: the executor composes the current object estimate with the relative stance observed in demonstrations, and refreshes this goal before every attempt and retry, so that a placement error does not carry over as a start-state error of the same size. Fixed devices use a task reference stance.

\textbf{2) Base station keeping.} During consecutive manipulations at one workstation the executor keeps a single world-frame stance reference, instead of re-anchoring per stage and accumulating error. A correction starts when the base drift exceeds an outer threshold and stops inside an inner one,
\begin{equation}
  \text{correct} \leftarrow \|p_B - p^{*}_B\| > d_{\mathrm{out}},\qquad
  \text{stop} \leftarrow \|p_B - p^{*}_B\| < d_{\mathrm{in}}.
  \label{eq:sk}
\end{equation}
The reason for masking rather than blending the VLA's base channel during manipulation is that a flat policy's base commands are the main source of drift in our trajectories.

\textbf{3) World-anchored contact targets.} For contacts whose tolerance is far below the base motion, such as a button, the target pose is stored in the world frame and re-expressed in the base frame at every control period,
\begin{equation}
  T^{B}_{E,d}(t) = \big(T^{W}_{B}(t)\big)^{-1}\, T^{W}_{E,d}(t),
  \label{eq:press}
\end{equation}
where $W$, $B$, and $E$ denote the world, base, and end-effector frames, and the updated target is converted to joint commands by inverse kinematics. The same rule underlies the stance reference in 2): what is locked is the target in the world, not the joint configuration, so the arm keeps aiming at the same point while the body sways. The same reference can also drive such a stage geometrically, which we evaluate as a separate variant.



%% file: experiments.tex

\section{Experimental Results}
\label{sec:exp}

\begin{table}[t]
\caption{Coffee fetching: stage progress and success (\%). Lift: mug lifted; Press: machine button triggered; Succ.: mug delivered after brewing. Avg.: row and column means.}
\label{tab:coffee}
\centering\footnotesize
\setlength{\tabcolsep}{3pt}
\resizebox{\columnwidth}{!}{
\begin{tabular}{lccccccc}
\toprule
 & \multicolumn{3}{c}{Nominal (C0)} & \multicolumn{3}{c}{Random init. (C1)} & \\
\cmidrule(lr){2-4}\cmidrule(lr){5-7}
Model & Lift & Press & Succ. & Lift & Press & Succ. & Avg. \\
\midrule
Flat VLA $\pi_{0.5}$ & 20 & 0 & 0 & 10 & 0 & 0 & 5.0 \\
FSM only $\pi_{0.5}$ & 92 & 48 & 32 & 95 & 55 & 50 & 62.0 \\
\method{} $\pi_{0.5}$ & \textbf{100} & 70 & 60 & \textbf{100} & 65 & 55 & 75.0 \\
\method{} + WCL-P $\pi_{0.5}$ & \textbf{100} & \textbf{98} & \textbf{70} & \textbf{100} & \textbf{95} & \textbf{70} & 88.8 \\
\midrule
Flat VLA GR00T & 70 & 35 & 35 & 40 & 25 & 20 & 37.5 \\
FSM only GR00T & 95 & 80 & 65 & 90 & 80 & 75 & 80.8 \\
\method{} GR00T & \textbf{100} & 85 & \textbf{80} & \textbf{100} & 90 & \textbf{80} & 89.2 \\
\method{} + WCL-P GR00T & \textbf{100} & \textbf{100} & \textbf{80} & \textbf{100} & \textbf{95} & 70 & 90.8 \\
\midrule
Avg. & 84.6 & 64.5 & 52.8 & 79.4 & 63.1 & 52.5 & 66.1 \\
\bottomrule
\end{tabular}}
\vspace{-3mm}
\end{table}

In this section, we analyze long-horizon task performance and stage progression, demonstrate the improvements of \method{} across multiple loco-manipulation tasks, evaluate the effectiveness of execution structure and stage-wise training, and conduct an ablation study of world-frame locking. We further validate \method{} on a real Unitree G1 humanoid robot.

\subsection{Experimental Setups}
\textbf{1) Environments and tasks.} Simulation experiments are performed in Isaac Lab Arena \cite{isaaclab-arena2025} with a Unitree G1 humanoid, Dex-3 hands, and the corresponding whole-body control stack; the actions of each model are mapped to a common executor interface. We choose three tasks covering bimanual transport, single-hand object transfer, and long-horizon manipulation with device interaction. \emph{Box transport}: take a box from a shelf and carry it to a target container, with 100 official MimicGen demonstrations (200 manipulation segments after cutting). \emph{Mug placement}: take a mug from a counter and transfer it to a target region, with 30 teleoperated demonstrations (60 segments). \emph{Coffee fetching}: take the mug, operate the coffee machine, re-grasp, and deliver, with 50 Quest-teleoperated demonstrations (250 segments); this is the main testbed. The real-robot task is to grasp a mug, place it in the machine region, re-grasp, and deliver, with 44 Quest demonstrations (176 segments). The real G1 uses the Dex-3 left hand; Vicon provides the poses of the robot, and head and wrist cameras provide policy observations. Simulation and real data (50\,Hz and 30\,Hz) are trained and deployed separately.

\textbf{2) Policies.} We use $\pi_{0.5}$ and GR00T N1.6 as skill backbones, fine-tuned on the same demonstrations. Both output upper-body joint targets, a base-height command, and a 3-D base-velocity command in chunks of 16 steps: GR00T N1.6 predicts a 35-D action (arm and hand joints, waist joints, and base commands), and $\pi_{0.5}$ a 32-D action without the waist, which the whole-body controller holds instead. The executor maps both to the same command interface, keeps the joint and base-height channels, and takes over the base-velocity channel during manipulation, so the execution structure is shared across backbones. Locomotion is provided by a learned whole-body controller that tracks these commands: HOMIE~\cite{ben2025homie} in simulation and the decoupled GR00T-WBC policy~\cite{nvidia2025grootwbc} on the real robot.

\begin{figure}[t]
  \centering
  \hspace{-4mm}
  \includegraphics[width=\columnwidth]{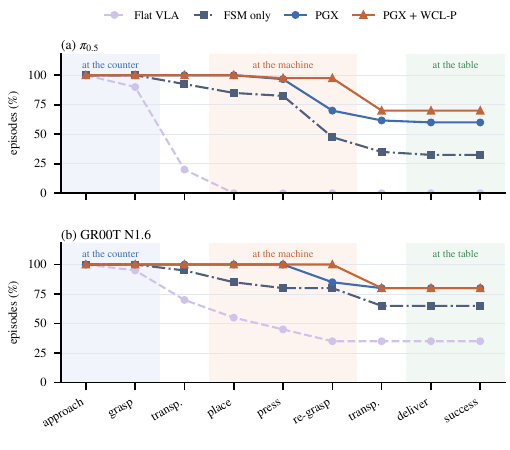}
  \vspace{-4mm}
  \caption{\textbf{Stage survival on coffee fetching} with (a) $\pi_{0.5}$ and (b) GR00T N1.6. Each line gives the share of episodes that reach a stage. We further visualize the difference in Fig.~\ref{fig:flat_frames}.}
  \label{fig:funnel}
\end{figure}

\begin{figure}[t]
  \centering
  \includegraphics[width=\columnwidth]{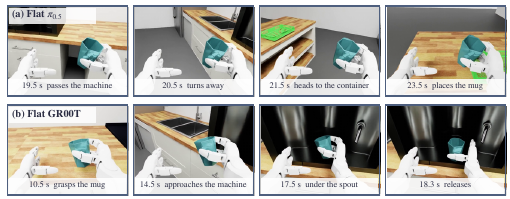}
\caption{\textbf{Where the flat policies differ.} Head-camera frames with episode time. (a) Flat $\pi_{0.5}$ passes the coffee machine with the mug and delivers it without brewing. (b) Flat GR00T reaches the machine and places the mug under the spout; its losses occur later, at the button and the re-grasp.}
  \label{fig:flat_frames}
  \vspace{-4mm}
\end{figure}

\textbf{3) Comparison methods.} To separate the contributions of execution structure, skill training, and spatial control, we compare four configurations: 
\begin{itemize}
    \item \emph{Flat VLA}: a flat model directly issues locomotion and manipulation commands;
    \item \emph{FSM only} (\method{} w/o WCL): stage gating and recovery are kept, but nominal stances are used, and object-relative stance and station keeping are switched off; 
    \item \emph{\method{}}: the full execution structure with the applicable WCL components, manipulation by the VLA; 
    \item \emph{\method{} + WCL-P}: the same, with the button stage driven
from the world-anchored primitive instead of the VLA.
\end{itemize}
On box transport and mug placement, \method{} uses the stage models of both backbones; on coffee fetching it uses the stage model for $\pi_{0.5}$ and the flat model for GR00T, and within one backbone the FSM-only and WCL-P variants keep the same skill model. Sec.~\ref{sec:exp-structure} reports both skill models to separate the executor from segmentation.


\begin{table}[t]
\caption{Stage progress and success on the three simulated tasks (\%). GR: GR00T. Lift: object lifted; Suc: task success. Nom.: nominal initial pose; Rand.: random initial pose.}
\label{tab:sim}
\centering\footnotesize
\setlength{\tabcolsep}{3.2pt}
\resizebox{\columnwidth}{!}{
\begin{tabular}{l cc cc cc cc cc cc c}
\toprule
 & \multicolumn{4}{c}{Box transport} & \multicolumn{4}{c}{Mug placement} & \multicolumn{4}{c}{Coffee fetching} & \\
\cmidrule(lr){2-5}\cmidrule(lr){6-9}\cmidrule(lr){10-13}
 & \multicolumn{2}{c}{Nom.} & \multicolumn{2}{c}{Rand.} & \multicolumn{2}{c}{Nom.} & \multicolumn{2}{c}{Rand.} & \multicolumn{2}{c}{Nom.} & \multicolumn{2}{c}{Rand.} & \\
\cmidrule(lr){2-3}\cmidrule(lr){4-5}\cmidrule(lr){6-7}\cmidrule(lr){8-9}\cmidrule(lr){10-11}\cmidrule(lr){12-13}
Model & Lift & Suc & Lift & Suc & Lift & Suc & Lift & Suc & Lift & Suc & Lift & Suc & Avg \\
\midrule
Flat GR & 95 & 55 & 25 & 20 & 35 & 15 & 5 & 5 & 70 & 35 & 40 & 20 & 35.0 \\
\method{} GR & \textbf{100} & 75 & 65 & 45 & 80 & 65 & 75 & 70 & \textbf{100} & \textbf{80} & \textbf{100} & \textbf{80} & 77.9 \\
Flat $\pi_{0.5}$ & 95 & 10 & 60 & 5 & 0 & 0 & 5 & 0 & 20 & 0 & 10 & 0 & 17.1 \\
\method{} $\pi_{0.5}$ & \textbf{100} & \textbf{90} & \textbf{75} & \textbf{60} & \textbf{90} & \textbf{75} & \textbf{90} & \textbf{90} & \textbf{100} & 60 & \textbf{100} & 55 & \textbf{82.1} \\
\midrule
Flat (avg) & 95.0 & 32.5 & 42.5 & 12.5 & 17.5 & 7.5 & 5.0 & 2.5 & 45.0 & 17.5 & 25.0 & 10.0 & 26.0 \\
\method{} (avg) & 100.0 & 82.5 & 70.0 & 52.5 & 85.0 & 70.0 & 82.5 & 80.0 & 100.0 & 70.0 & 100.0 & 67.5 & 80.0 \\
\bottomrule
\end{tabular}
}
\vspace{-2mm}
\end{table}

\textbf{4) Conditions and metrics.} Each task is evaluated under a nominal initial pose (C0) and a randomized initial pose (C1) that adds uniform offsets of $\pm0.30$\,m in both translation axes and $\pm30^{\circ}$ in yaw; demonstrations were all recorded from the nominal pose, so C1 is a start the flat policies have never seen. Comparisons within a task share the time budget (40\,s, 60\,s, and 110\,s for box, mug, and coffee) and the success criterion, with at least 20 evaluation episodes per cell; coffee requires the device to be triggered and the mug finally placed and released, so delivering the mug directly does not count. 
We record end-to-end success, per-episode stage progress (the farthest stage reached, to attribute end-to-end losses), and base drift during manipulation.

\subsection{Long-Horizon Task: Coffee Fetching}
\label{sec:exp-coffee}

Table~\ref{tab:coffee} reports stage progress and success on coffee fetching, and Fig.~\ref{fig:funnel} shows, per stage, the share of episodes that survive. \method{} lifts $\pi_{0.5}$ from 0\% under both initial conditions to 60\% and 55\%, and GR00T from 35\% / 20\% to 80\% / 80\%. The GR00T gain is obtained with unchanged policy weights, indicating that the manipulation ability a VLA already has can be turned into long-horizon success by organizing its execution. The FSM-only rows separate the two contributions: the state machine alone already lifts the mug in 90--95\% of the episodes, and adding world-frame locking raises the press rate further (to 65--70\% for $\pi_{0.5}$ and 85--90\% for GR00T) and the end-to-end success by up to 28 points.


The flat $\pi_{0.5}$ policy never completes the machine interaction and in several episodes skips the device to deliver the mug directly; \method{} makes the dependency explicit, so a later stage is entered only when its physical precondition holds. Executing the button stage with the world-anchored primitive (WCL-P) raises the press rate to 95--100\% for both backbones and the success of $\pi_{0.5}$ to 70\%, confirming that what remains at the button is a matter of millimeter precision rather than of sequencing; GR00T presses the button reliably already, and its remaining losses lie at the re-grasp.

The funnels make the attribution visible. With the flat policies the losses are spread over the early stages (GR00T) or concentrated before the machine ($\pi_{0.5}$, which skips the machine, Fig.~\ref{fig:flat_frames}); inside \method{}, every episode reaches the machine, and once the primitive handles the press, the main remaining loss is the re-grasp from the machine tray, a property of the skill rather than the execution.

\subsection{Effectiveness on Multiple Loco-Manipulation Tasks}
Table~\ref{tab:sim} reports stage progress and task success under both conditions. On all three tasks, \method{} lifts both backbones from near failure to reliable execution: across the six task-condition settings, the average of stage progress and success rises from 35\% to 78\% for GR00T and from 17\% to 82\% for $\pi_{0.5}$. The failure modes differ across tasks: on box transport the flat policies lift the box but lose it on the way, whereas on coffee fetching the flat $\pi_{0.5}$ passes the machine.



\begin{table}[t]
\caption{Execution structure versus skill training on coffee fetching success (\%). Avg: row and column means.}
\label{tab:structure}
\centering\footnotesize
\setlength{\tabcolsep}{5pt}
\begin{tabular}{lcccc}
\toprule
Backbone / cond. & Flat & \method{} flat & \method{} stage & Avg. \\
\midrule
$\pi_{0.5}$ C0 & 0 & \textbf{60} & \textbf{60} & 40.0 \\
$\pi_{0.5}$ C1 & 0 & \textbf{60} & 55 & 38.3 \\
\midrule
GR00T C0 & 35 & \textbf{80} & 35 & 50.0 \\
GR00T C1 & 20 & \textbf{80} & 35 & 45.0 \\
\midrule
Avg. & 13.8 & \textbf{70.0} & 46.3 & 43.3 \\
\bottomrule
\end{tabular}
\vspace{-1mm}
\end{table}

\begin{figure}[t]
  \centering
\includegraphics[width=\columnwidth]{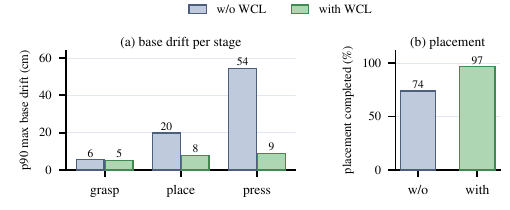}
  \vspace{-6mm}
  \caption{\textbf{Station keeping under sustained base disturbance} ($\pi_{0.5}$, C0). (a) p90 of the per-stage maximum base drift with and without WCL. (b) Completed placement attempts.}
  \label{fig:drift}
  \vspace{-4mm}
\end{figure}



On mug placement, the flat policies rarely get past the grasp: GR00T lifts the mug in 35\% of the nominal episodes and places it in 15\%, and $\pi_{0.5}$ in none. 
Inside \method{}, GR00T reaches 65\% and 70\% under the nominal and random poses, and $\pi_{0.5}$ reaches 75\% and 90\%, showing improved task completion with both backbones.
The random initial pose costs \method{} nothing here, because the object-relative stance and station keeping re-establish the same start state before every grasp.

Box transport uses the official demonstrations, and the same executor applies despite the different task structure. The lift column separates arriving at the box from carrying it: under the nominal pose both flat policies lift the box in 95\% of the episodes but deliver it in only 55\% (GR00T) and 10\% ($\pi_{0.5}$), because the box is lost during transport; under the random initial pose the lift rate falls to 25\% and 60\%. \method{} lifts the box in every nominal episode and delivers it in 75\% and 90\%, and under the random pose it more than doubles the flat success rates, showing that navigating to a standard manipulation stance absorbs the change of the starting point before it reaches the skill.

\begin{figure}[t]
  \centering
  \includegraphics[width=0.75\columnwidth]{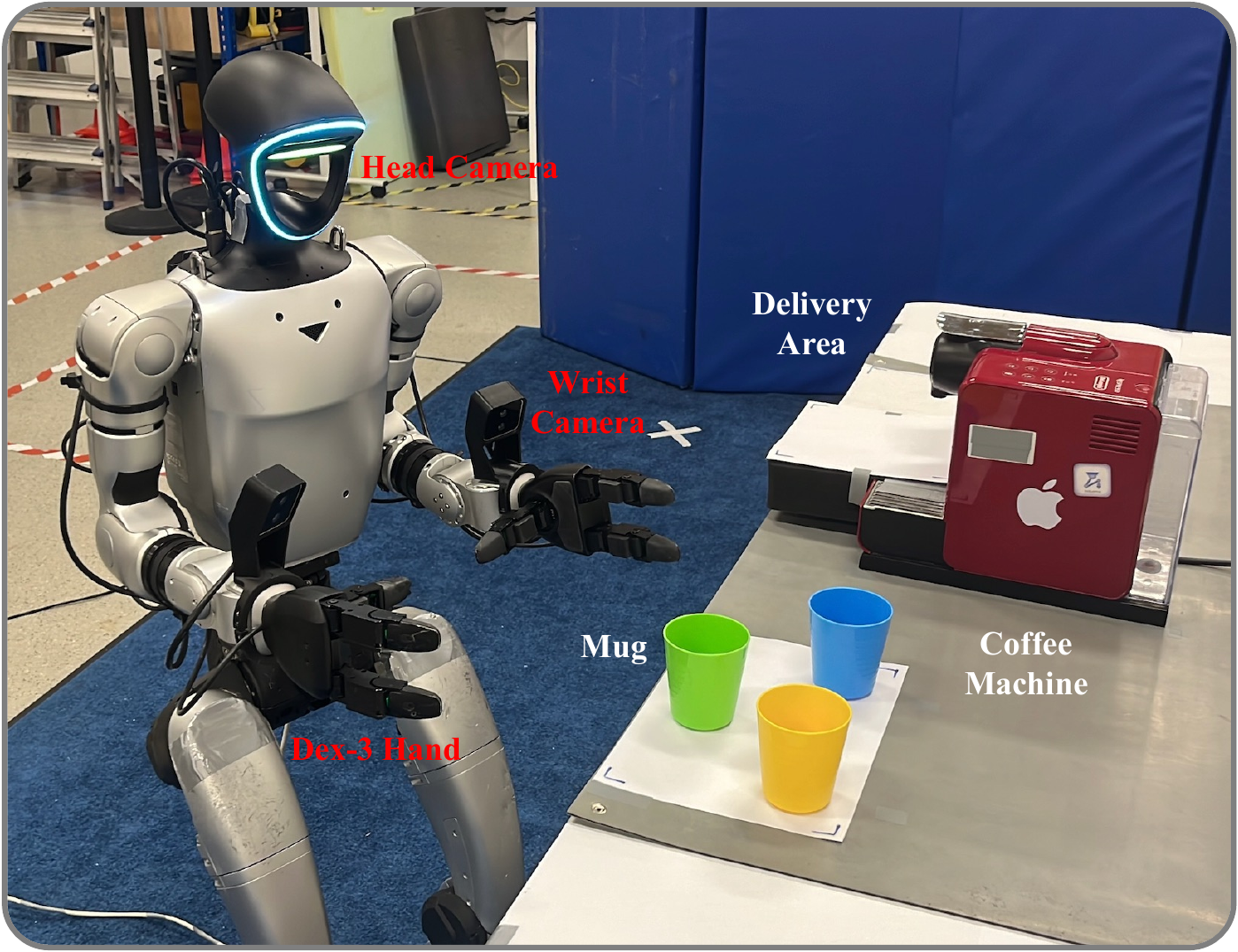}
  \caption{\textbf{Real-robot setup.} The Unitree G1 with Dex-3 hands and head and wrist cameras at the workstation with the mug, the coffee machine, and the delivery area.}
  \label{fig:real}
  \vspace{-2mm}
\end{figure}

\begin{figure}[t]
  \centering
\hspace{-4mm}  \includegraphics[width=0.8\columnwidth]{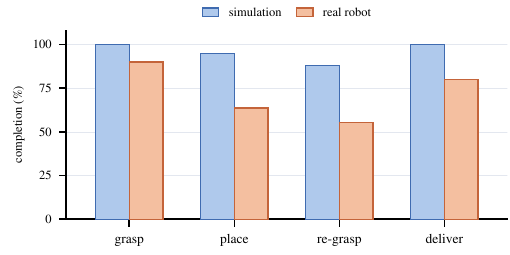}
  \caption{\textbf{Per-stage completion on the real G1} compared with the same stages in simulation (\method{} $\pi_{0.5}$ stage model, C0).}
  \label{fig:real_stages}
  \vspace{-4mm}
\end{figure}

\subsection{Execution Structure versus Stage-Wise Training}
\label{sec:exp-structure}
This section and Sec.~\ref{sec:exp-wcl} form the ablation of the executor: the ladder flat $\rightarrow$ + execution structure $\rightarrow$ + spatial stabilization, crossed with the two skill models. To isolate the executor itself, we place the flat model inside \method{} and keep the same task-level instruction in every manipulation stage. The first two columns of Table~\ref{tab:structure} therefore hold the weights and the language input fixed and change only the execution; the third column replaces the model by the stage-trained one.



For both backbones the executor brings a large gain without retraining the policy. Stage-wise training, in contrast, is a secondary and model-dependent factor: for $\pi_{0.5}$ it changes success by at most 5 points, and for GR00T the flat weights are the stronger skill model on coffee. \method{} therefore separates execution structure from skill weights, so that whichever model has the stronger local skill can be used directly, without retraining a stage model.

On the $\pi_{0.5}$ training frames we inspected, replacing the stage instruction by a mismatched one changes the output little. Together with the same-weight ablation, this indicates that stage prompts alone do not explain the gain of \method{}: segmentation changes the distribution of manipulation data, whereas the executor provides the stage organization and start-state control. This also suggests that a planner that only switches the prompt would not by itself recover this gain.

\subsection{Effect of World-Frame Stabilization}
\label{sec:exp-wcl}

\textbf{Stance and drift.} Switching off spatial stabilization (the FSM-only rows of Table~\ref{tab:coffee}) drops $\pi_{0.5}$ on coffee from 60\% / 55\% to 32\% / 50\% and GR00T from 80\% / 80\% to 65\% / 75\%. This comparison removes object-relative stance and station keeping together, and reflects their joint contribution. To observe drift within a stage, we inject a base displacement in a fixed direction every 3\,s during manipulation. Fig.~\ref{fig:drift} shows that station keeping bounds the accumulated drift and raises the completion ratio of placement attempts.


Without injected disturbance the natural drift of a standing humanoid is 3--4\,cm, within the station-keeping band, so the benefit comes from the object-relative stance that re-establishes the start state before each skill; under sustained disturbance the error would otherwise accumulate across consecutive manipulations, and station keeping 
reduces the p90 of per-stage maximum drift to 9\,cm, compared with 54\,cm without it.
Millimeter-level contact under such disturbance additionally requires the base to come to rest, which the executor enforces before the button stage. The predicate interface itself tolerates state noise: on the mug task, per-episode object-position biases of 2, 3, and 5\,cm with per-frame jitter keep success at 75--90\%.



\begin{figure}[t]
\vspace{-1mm}
  \centering  \includegraphics[width=\columnwidth]{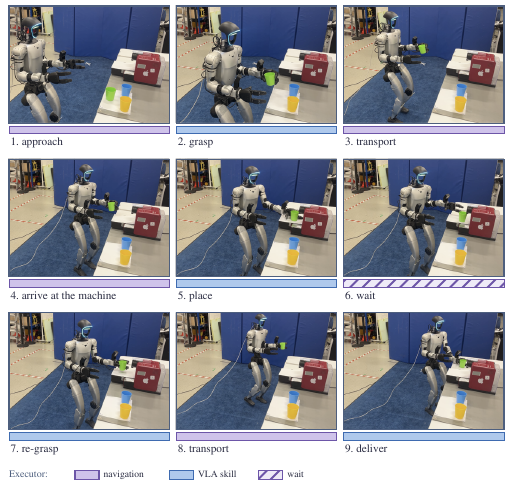}
  \caption{\textbf{Real-robot coffee task on the Unitree G1.} External-camera frames of one episode, with the executor of each stage marked below the frame. Predicates read Vicon and joint states; the executor and its stage definitions are transferred from simulation.}
  \label{fig:real_frames}
  \vspace{-4mm}
\end{figure}

\textbf{Fine pressing.} Over 14 evaluation runs without sustained disturbance (two backbones, two skill models, and primitive versions), the world-anchored primitive completed 243 of 245 entries into the button stage (99\%), against 72--100\% when the stage is left to the VLA inside the full \method{}, both conditioned on reaching the stage.


\subsection{Real-Robot Experiments}
\label{sec:exp-real}
Fig.~\ref{fig:real} shows our real experiment setup.
We deploy \method{} on a G1 with a transferred stage graph and predicates from simulation; the predicates read motion capture and joint states, and the VLA is fine-tuned on 44 real demonstrations. The task keeps the structure of the simulated one: the robot walks to the counter, grasps the mug, carries it to the coffee machine, places it, waits, re-grasps, and delivers it (Fig.~\ref{fig:real_frames}).

Fig.~\ref{fig:real_stages} reports the completion rates of the four manipulation stages, all executed by the PGX $\pi_{0.5}$ stage model; each rate is computed over the episodes that reach that stage. Grasping transfers best among these stages, and the differences from simulation concentrate at the two contact-rich stages, placing and re-grasping. These gaps reflect the precision of the fine-tuned skills on the real hand rather than the execution structure.



%% file: conclusion.tex

\section{Conclusion}
In this work, we study long-horizon loco-manipulation on a humanoid robot and propose \method{}, which composes existing VLA skills into a long task through predicate-driven stage progression, skill execution under controlled start conditions, and world-frame spatial stabilization. Experiments on three simulated tasks and two backbones show that explicit execution structure raises task completion substantially, while stage-wise training and precision control contribute at different levels. We also validate our method on a real G1 humanoid. \method{} thus provides an effective and general framework bridging robot foundation models and classical control for loco-manipulation tasks. 

\textbf{Limitation and Future Work.} \method{} achieves higher success rates yet costs longer execution time for stage transitions, which we will improve in future work, together with extending the framework to richer skills and tasks.
\vspace{6mm}
